\documentclass[a4paper,conference]{IEEEtran}
\usepackage{amsmath,epsfig}
\usepackage{cite}
\usepackage{framed,multirow,booktabs}
\usepackage{graphicx}
\usepackage{amssymb,bm,amsfonts}
\usepackage{algorithmic}
\usepackage{textcomp}
\usepackage{amsthm}
\usepackage{lineno}
\ifCLASSINFOpdf
\else
\fi

\begin{document}
%
\title{Multi-Scale Attention with Dense Encoder for Handwritten Mathematical Expression Recognition}

\author{\IEEEauthorblockN{Jianshu Zhang,
Jun Du and
Lirong Dai}
\IEEEauthorblockA{National Engineering Laboratory for Speech and Language Information Processing\\
University of Science and Technology of China,
Hefei, Anhui, P. R. China\\ Email: xysszjs@mail.ustc.edu.cn, jundu@ustc.edu.cn, lrdai@ustc.edu.cn}
}

\maketitle

\begin{abstract}
  Handwritten mathematical expression recognition is a challenging problem due to the complicated two-dimensional structures, ambiguous handwriting input and variant scales of handwritten math symbols. To settle this problem, recently we propose the attention based encoder-decoder model that recognizes mathematical expression images from two-dimensional layouts to one-dimensional LaTeX strings. In this study, we improve the encoder by employing densely connected convolutional networks as they can strengthen feature extraction and facilitate gradient propagation especially on a small training set. We also present a novel multi-scale attention model which is employed to deal with the recognition of math symbols in different scales and restore the fine-grained details dropped by pooling operations. Validated on the CROHME competition task, the proposed method significantly outperforms the state-of-the-art methods with an expression recognition accuracy of 52.8\% on CROHME 2014 and 50.1\% on CROHME 2016, by only using the official training dataset.
\end{abstract}


\IEEEpeerreviewmaketitle

\section{Introduction}
\label{sec:Introduction}
Mathematical expressions are indispensable for describing problems in maths, physics and many other fields. Meanwhile, people have begun to use handwritten mathematical expressions as one natural input mode. However, machine recognition of these handwritten mathematical expressions is difficult and exhibits three distinct challenges~\cite{belaid1984syntactic}, i.e., the complicated two-dimensional structures, enormous ambiguities coming from handwriting input and variant scales of handwritten math symbols.

Handwritten mathematical expression recognition comprises two major problems~\cite{chan2000mathematical}: symbol recognition and structural analysis. The two problems can be solved sequentially~\cite{zanibbi2002recognizing} or globally~\cite{alvaro2016integrated}. However, both conventional sequential and global approaches have the following limitations: 1) the challenging symbol segmentation is inevitable, which brings many difficulties; 2) the structural analysis is commonly based on two-dimensional context free grammar~\cite{chou1989recognition}, which requires priori knowledge to define a math grammar; 3) the complexity of parsing algorithms increases with the size of math grammar.

In recent research of deep learning, a novel attention based encoder-decoder model has been proposed~\cite{bahdanau2014neural,sutskever2014sequence}. Its general application in machine translation~\cite{cho2014learning}, speech recognition~\cite{bahdanau2016end}, character recognition~\cite{zhang2017ran,zhang2017trajectory} and image captioning~\cite{xu2015show} inspires researchers that mathematical expression recognition can also be one proper application~\cite{zhang2017watch,zhang2017gru,deng2016you,anh2017training}. More specifically, \cite{zhang2017watch} proposed a model namely WAP. The WAP learns to encode input expression images and decode them into LaTeX strings. The encoder is a convolutional neural network (CNN)~\cite{krizhevsky2012imagenet} based on VGG architecture~\cite{simonyan2014very} that maps images to high-level features. The decoder is a recurrent neural network (RNN)~\cite{graves2013speech} with gated recurrent units (GRU)~\cite{chung2014empirical} that converts these high-level features into output strings one symbol at a time. For each predicted symbol, an attention model built in the decoder scans the entire input expression image and chooses the most relevant region to describe a math symbol or an implicit spatial operator. Compared with conventional approaches for handwritten mathematical expression recognition, the attention based encoder-decoder model possesses three distinctive properties: 1) It is end-to-end trainable; 2) It is data-driven, in contrast to traditional systems that require a predefined math grammar; 3) Symbol segmentation can be automatically performed through attention model.

\begin{figure}
\centering
\includegraphics[width=3in]{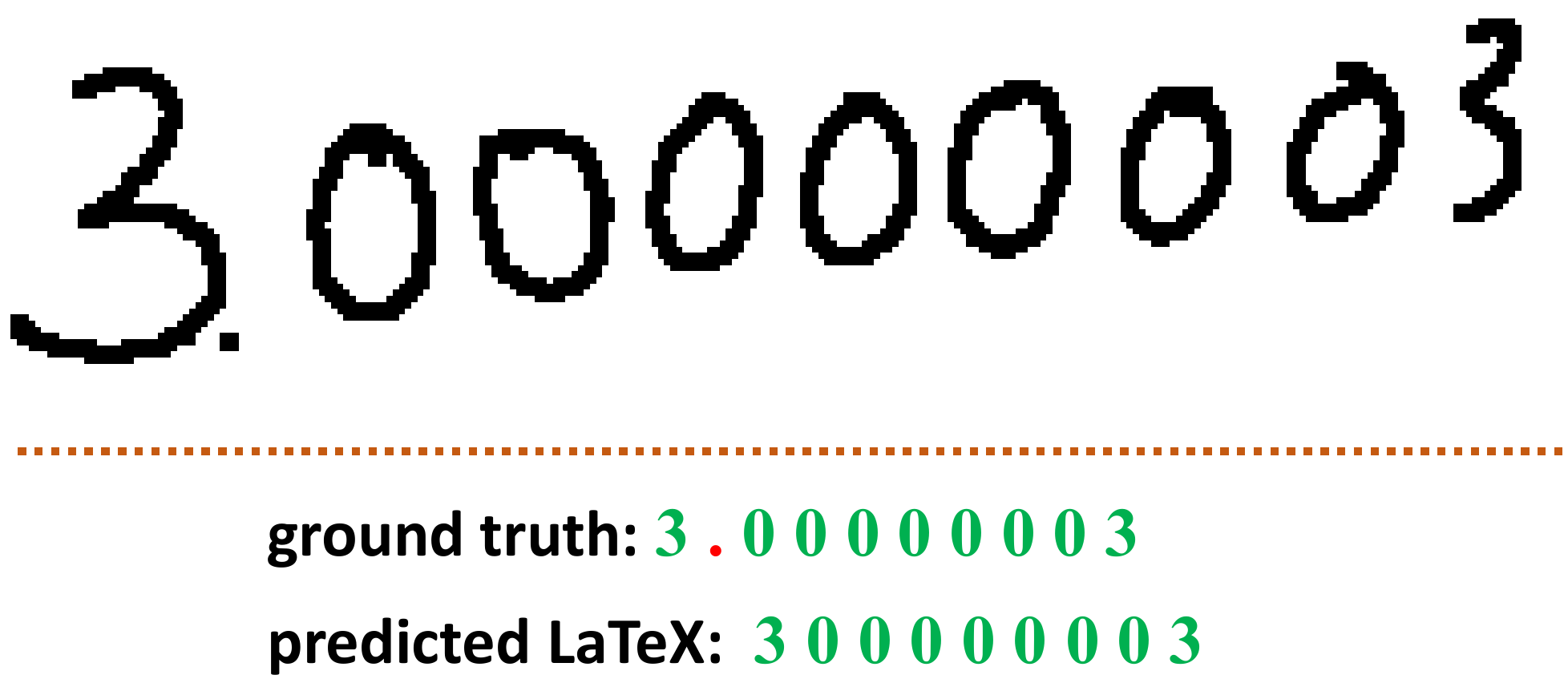}
\caption{An incorrectly recognized example of handwritten mathematical expression due to the under-parsing problem: the decimal point ``.'' is missed in the predicted LaTeX notation.}
\label{fig:under-parsing-example}
\end{figure}

In this study, we still focus on offline handwritten mathematical expression recognition and report our recent progress on WAP model~\cite{zhang2017watch}. The main contribution is in two aspects. Firstly, we improve the CNN encoder by employing a novel architecture called densely connected convolutional networks (DenseNet)~\cite{huang2016densely}. The DenseNet has shown excellent performance on image classification task as it strengthens feature extraction and facilitates gradient propagation. Secondly, we present a novel multi-scale attention model to deal with the problems caused by pooling operations. Although pooling layers are essential parts of convolutional networks, they shrink the size of feature maps, yielding decrease of resolution. Because the scales of handwritten math symbols vary severely, the fine-grained details of extracted feature maps are especially important in handwritten mathematical expression recognition, which are lost in low-resolution feature maps. For example, in Fig.~\ref{fig:under-parsing-example}, the decimal point is very close to math symbol ``3'' and its scale is much smaller than its adjacent symbols. After many pooling layers, the visual information of the decimal point is gone, which leads to an under-parsing problem. To implement the multi-scale attention model, we propose a multi-scale dense encoder that will provide both low-resolution features and high-resolution features. The low-resolution features capture a larger receptive field and are more semantic while the high-resolution features restore more fine-grained visual information. The decoder then attends to both low-resolution and high-resolution features for predicting output LaTeX strings.

The remainder of the paper is organized as follows. In Section~\ref{sec:Methodology}, we introduce the dense encoder and the proposed multi-scale attention model in detail. We introduce the implementation of training and testing procedure in Section~\ref{sec:Training and Testing Details}.
The performances of dense encoder and multi-scale attention model are shown through experimental results and visualization analysis in Section~\ref{sec:Experiments}. Finally we conclude this study in Section~\ref{sec:Conclusion}.

\section{Methodology}
\label{sec:Methodology}
In this section, we first make a brief summarization of DenseNet since our encoder is based on densely connected convolutional blocks. Then we introduce the classic attention based encoder-decoder framework. Finally, we extend DenseNet by introducing a multi-scale dense encoder and describe the implementation of multi-scale attention model in detail.

\subsection{Dense Encoder}
\label{sec:Dense Encoder}
The main idea of DenseNet is to use the concatenation of output feature maps of preceding layers as the input of succeeding layers. As DenseNet is composed of many convolution layers, let $H_{l}(\cdot)$ denote the convolution function of the $l^{\textrm{th}}$ layer, then the output of layer $l$ is represented as:
\begin{equation}\label{eq:dense output}
\mathbf{x}_{l}=H_{l}([\mathbf{x}_{0};\mathbf{x}_{1};\ldots;\mathbf{x}_{l-1}])
\end{equation}
where $\mathbf{x}_{0}, \mathbf{x}_{1},\ldots, \mathbf{x}_{l}$ denote the output features produced in layers $0, 1, \ldots, l$, ``$;$'' denotes the concatenation operation of feature maps. This iterative connection enables the network to learn shorter interactions cross different layers and reuses features computed in preceding layers. By doing so the DenseNet strengthens feature extraction and facilitates gradient propagation.

An essential part of convolutional networks is pooling layers, which is capable of increasing receptive field and improving invariance. However, the pooling layers disenable the concatenation operation as the size of feature maps changes. Also, DenseNet is inherently memory demanding because the number of inter-layer connections grows quadratically with depth. Consequently, the DenseNet is divided into multiple densely connected blocks as shown in Fig.~\ref{fig:multi-scale-dense}. A compression layer is appended before each pooling layer to further improve model compactness.

\subsection{Decoder}
\label{sec:Decoder with Attention}
We employ GRU as the decoder because it is an improved version of simple RNN which can alleviate the vanishing and exploding gradient problems~\cite{bengio1994learning,zhang2016rnn}. Given input ${{\bf{x}}_t}$, the GRU output ${\mathbf{h}_t}$ is computed by:
\begin{equation}\label{eq:GRU function}
{{\bf{h}}_t} = \textrm{GRU} \left( {{\bf{x}}_t}, {{\bf{h}}_{t - 1}} \right)
\end{equation}
and the GRU function can be expanded as follows:
\begin{align}\label{eq:expandGRU}
 & {{\mathbf{z}}_t} = \sigma ({{\mathbf{W}}_{xz}}{{\mathbf{x}}_{t}} + {{\mathbf{U}}_{hz}}{{\mathbf{h}}_{t - 1}}) \\
 & {{\mathbf{r}}_t} = \sigma ({{\mathbf{W}}_{xr}}{{\mathbf{x}}_{t}} + {{\mathbf{U}}_{hr}}{{\mathbf{h}}_{t - 1}}) \\
 & {{\bf{\tilde h}}_t} = \tanh ({{\bf{W}}_{xh}}{{\bf{x}}_{t}} + {{\bf{U}}_{rh}}({{\bf{r}}_t} \otimes {{\bf{h}}_{t - 1}})) \\
 & {{\bf{h}}_t} = (1 - {{\bf{z}}_t}) \otimes {{\bf{h}}_{t - 1}} + {{\bf{z}}_t} \otimes {{\bf{\tilde h}}_t}
\end{align}
where $\sigma$ is the sigmoid function and $\otimes$ is an element-wise multiplication operator. ${{\mathbf{z}}_t}$, ${{\mathbf{r}}_t}$ and ${{\bf{\tilde h}}_t}$ are the update gate, reset gate and candidate activation, respectively.

Assuming the output of CNN encoder is a three-dimensional array of size $H \times W \times C$, consider the output as a variable-length grid of $L$ elements, $L=H \times W$. Each of these elements is a $C$-dimensional annotation that corresponds to a local region of the image.
\begin{equation}\label{eq:annotations}
 \mathbf{A} = \left\{ { \mathbf{a}_1, \ldots ,\mathbf{a}_L} \right\}\;,\;{{\mathbf{a}}_i} \in {\mathbb{R}^C}
\end{equation}
Meanwhile, the GRU decoder is employed to generate a corresponding LaTeX string of the input mathematical expression. The output string $\mathbf{Y}$ is represented by a sequence of one-hot encoded symbols.
\begin{equation}\label{eq:outputY}
 \mathbf{Y} = \left\{ { \mathbf{y}_1, \ldots ,\mathbf{y}_T} \right\}\;,\;{{\mathbf{y}}_i} \in {\mathbb{R}^K}
\end{equation}
where $K$ is the number of total symbols in the vocabulary and $T$ is the length of LaTeX string.

Note that, both the annotation sequence $\mathbf{A}$ and the LaTeX string $\mathbf{Y}$ are not fixed-length. To address the learning problem of variable-length annotation sequences and associate them with variable-length output sequences, we attempt to compute an intermediate fixed-length vector ${{\mathbf{c}}_t}$, namely context vector, at each decoding step $t$. The context vector ${{\mathbf{c}}_t}$ is computed via weighted summing the variable-length annotations ${{\mathbf{a}}_i}$:
\begin{equation}\label{eq:context vector}
  {{\mathbf{c}}_t} = \sum\nolimits_{i=1}^L {{\alpha _{ti}}{{\mathbf{a}}_i}}
\end{equation}
Here, the weighting coefficients $\alpha _{ti}$ are called attention probabilities and they will make decoder to know which part of input image is the suitable place to attend to generate the next predicted symbol and then assign a higher weight to the corresponding local annotation vectors ${{\mathbf{a}}_i}$.
After computing the intermediate fixed-length context vector, we then generate the LaTeX string one symbol at a time. By doing so, the problem of associating variable-length annotation sequences with variable-length output LaTeX strings is addressed.

The probability of each predicted symbol is computed by the context vector ${{\mathbf{c}}_t}$, current decoder state ${{\mathbf{s}}_t}$ and previous target symbol ${{\mathbf{y}}_{t - 1}}$ using the following equation:
\begin{equation}\label{eq:computePy}
  p({{\mathbf{y}}_t}|{{{\mathbf{y}}_{t - 1}},\mathbf{X}}) = g \left ({{\mathbf{W}}_o}h({\mathbf{E}}{{\mathbf{y}}_{t - 1}} + {{\mathbf{W}}_s}{{\mathbf{s}}_t} + {{\mathbf{W}}_c}{{\mathbf{c}}_t})\right )
\end{equation}
where $\mathbf{X}$ denotes input mathematical expression images, $g$ denotes a softmax activation function, $h$ denotes a maxout activation function, let $m$ and $n$ denote the dimensions of embedding and GRU decoder state respectively, then ${{\mathbf{W}}_o} \in {\mathbb{R}^{K \times \frac{m}{2}}}$ and ${{\mathbf{W}}_s} \in {\mathbb{R}^{m \times n}}$, 
${\mathbf{E}}$ denotes the embedding matrix.

\subsection{Multi-Scale Attention with Dense Encoder}
\label{sec:Multi-Scale Attention with Dense Encoder}

\subsubsection{Multi-Scale Dense Encoder}
\label{sec:Multi-Scale Dense Encoder}

\begin{figure}
\centering
\includegraphics[width=3.25in]{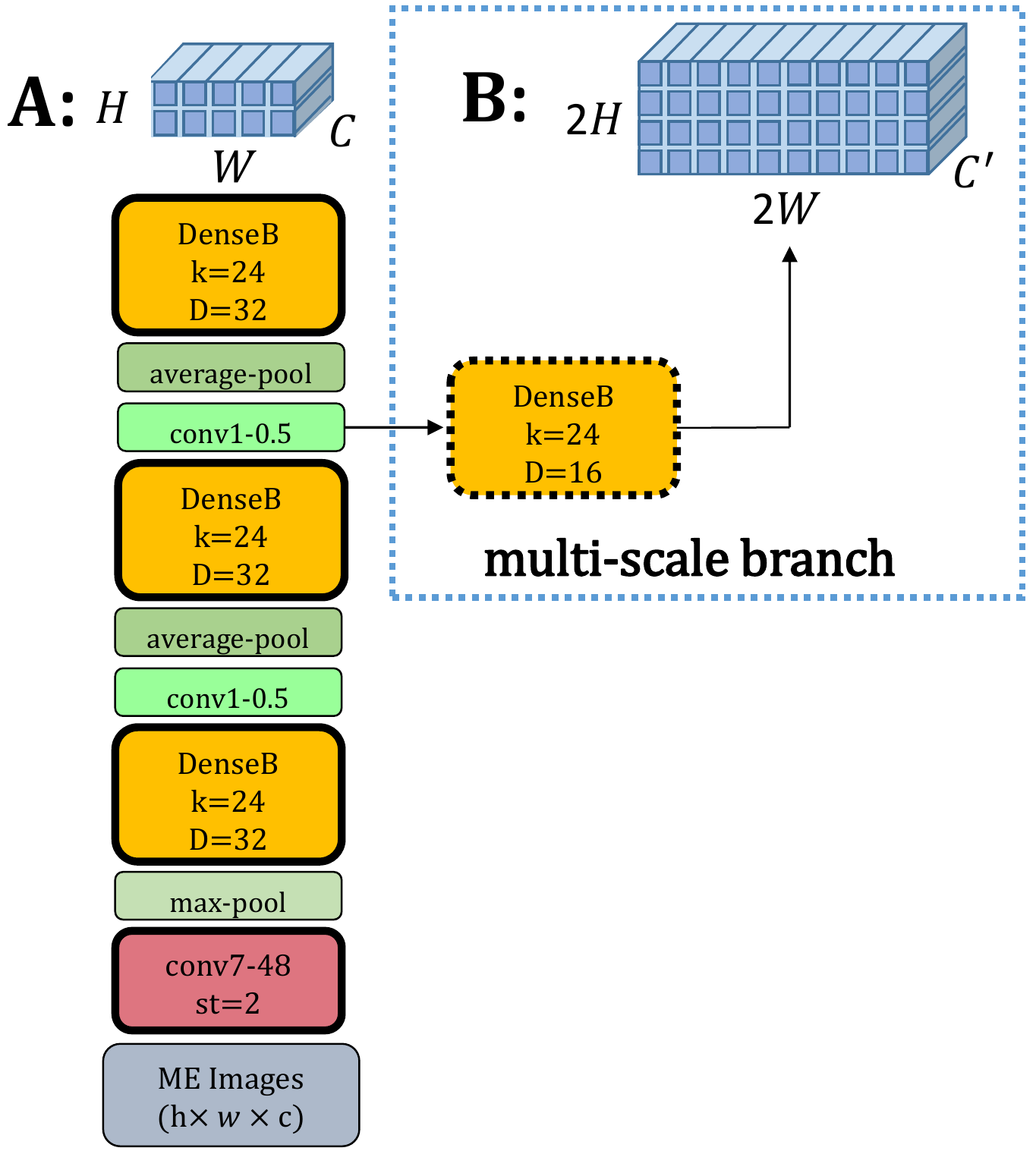}
\caption{Architecture of multi-scale dense encoder. The left part is the main branch while the right part is the multi-scale branch.}
\label{fig:multi-scale-dense}
\end{figure}

To implement the multi-scale attention model, we first extend the single-scale dense encoder into multi-scale dense encoder. As illustrated in Fig.~\ref{fig:multi-scale-dense}, our dense encoder consists of two branches, i.e., except the main branch which produces low-resolution annotations $\mathbf{A}$, our dense encoder has another multi-scale branch that produces high-resolution annotations $\mathbf{B}$. The multi-scale branch is extended before the last pooling layer of the main branch so that the output feature maps of multi-scale branch has a higher resolution. The high-resolution annotation is a three-dimensional array of size $2H \times 2W \times {C^{'}}$, which can be represented as a variable-length grid of $4L$ elements:
\begin{equation}\label{eq:new annotations}
 \mathbf{B} = \left\{ { \mathbf{b}_1, \ldots ,\mathbf{b}_{4L}} \right\}\;,\;{{\mathbf{b}}_i} \in {\mathbb{R}^{C^{'}}}
\end{equation}
where $L$ is the length of annotation sequence $A$. Intuitively, we can extend several multi-scale branches before every pooling layer but such operation brings too much computational cost as the size of feature maps becomes too large.

As for the implementation details of dense encoder, we employ three dense blocks in the main branch as described by yellow rectangles in Fig.~\ref{fig:multi-scale-dense}. Before entering the first dense block, a $7 \times 7$ convolution (stride is $2 \times 2$) with 48 output channels is performed on the input expression images, followed by a $2 \times 2$ max pooling layer. Each dense block is titled as ``DenseB'' because we use bottleneck layers to improve computational efficiency, i.e. a $1 \times 1$ convolution is introduced before each $3 \times 3$ convolution to reduce the input to $4k$ feature maps.
The growth rate $k=24$ and the depth (number of convolution layers) of each block $D=32$ which means each block has 16 $1 \times 1$ convolution layers and 16 $3 \times 3$ convolution layers. A batch normalization layer~\cite{ioffe2015batch} and a ReLU activation layer~\cite{glorot2011deep} are performed after each convolution layer consecutively. We use $1 \times 1$ convolution followed by $2 \times 2$ average pooling as transition layers between two contiguous dense blocks. The transition layer reduces the number of feature maps of each block by half. While in the multi-scale branch, we append another dense block with bottleneck layer, $k=24$ and $D=16$. We investigate the depth of block in multi-scale branch ($D=0, 8, 16, 24$) in Section~\ref{sec:Experiments}.

\subsubsection{Multi-Scale Attention Model}
\label{sec:Multi-Scale Attention Model}
In this study, our decoder adopts two unidirectional GRU layers to calculate the decoder state ${\mathbf{s}_t}$ and the multi-scale context vector ${\mathbf{c}_t}$ that are both used as input to calculate the probability of predicted symbol in Eq.~\eqref{eq:computePy}. We employ two different single-scale coverage based attention model to generate the low-resolution context vector and high-resolution context vector by attending to low-resolution annotations and high-resolution annotations respectively. As the low-resolution context vector and high-resolution context vector have the same length $1$, we concatenate them to produce the multi-scale context vector:
\begin{align}\label{eq:compute decoder state}
 & {{\mathbf{\hat s}}_t} = \textrm{GRU} \left( {{\bf{y}}_{t-1}}, {{\bf{s}}_{t - 1}} \right) \\
 & {\mathbf{cA}_t} = f_{\text{catt}} \left( \mathbf{A}, \mathbf{\hat s}_t \right) \\
 & {\mathbf{cB}_t} = f_{\text{catt}} \left( \mathbf{B}, \mathbf{\hat s}_t \right) \\
 & {\mathbf{c}_t} = [{\mathbf{cA}_t};{\mathbf{cB}_t}] \\
 & {{\mathbf{s}}_t} = \textrm{GRU} \left( {{\mathbf{c}}_t}, {{\mathbf{\hat s}}_t} \right)
\end{align}
where ${{\mathbf{s}}_{t-1}}$ denotes the previous decoder state, ${{\mathbf{\hat s}}_t}$ is the prediction of current decoder state, ${\mathbf{cA}_t}$ is the low-resolution context vector at decoding step $t$, similarly ${\mathbf{cB}_t}$ is the high-resolution context vector. The multi-scale context vector ${\mathbf{c}_t}$ is the concatenation of ${\mathbf{cA}_t}$ and ${\mathbf{cB}_t}$ and it performs as the input during the computation of current decoder state ${{\mathbf{s}}_{t}}$.

$f_{\text{catt}}$ denotes a single-scale coverage based attention model. Take the computation of low-resolution context vector ${\mathbf{cA}_t}$ as an example, we parameterize $f_{\text{catt}}$ as a multi-layer perceptron:
\begin{align}\label{eq:coverage attention}
  & {\mathbf{F}} = {\mathbf{Q}} * \sum\nolimits_{l=1}^{t - 1} {{{\bm{\alpha}}_l}} \\
  & {e_{ti}} = {\bm{\nu }}_{\text{att}}^{\rm T}\tanh ({{\mathbf{U}}_{s}}{{\mathbf{\hat s}}_t} + {{\mathbf{U}}_a}{{\mathbf{a}}_i} + {{\mathbf{U}}_f}{{\mathbf{f}}_i}) \\
  & {\alpha _{ti}} = \frac{{\exp ({e_{ti}})}}{{\sum\nolimits_{k = 1}^L {\exp ({e_{tk}})} }} \\
  & {{\mathbf{cA}}_t} = \sum\nolimits_{i=1}^L {{\alpha _{ti}}{{\mathbf{a}}_i}}
\end{align}
where ${{\mathbf{a}}_{i}}$ denotes the element of low-resolution annotation sequence $\mathbf{A}$,
${e_{ti}}$ denotes the energy of ${{\mathbf{a}}_{i}}$ at time step $t$ conditioned on the prediction of current decoder state ${{\mathbf{\hat s}}_t}$ and coverage vector ${{\mathbf{f}}_i}$. The coverage vector is initialized as a zero vector and we compute it based on the summation of all past attention probabilities. Hence the coverage vector contains the information of alignment history. We append the coverage vector in the attention model so that the decoder is capable to know which part of input image has been attended or not~\cite{zhang2017watch,tu2016modeling}. Let $n'$ denote the attention dimension and $q$ denote the number of output channels of convolution function $\mathbf{Q}$; then ${{\bm{\nu }}_{\text{att}}} \in {\mathbb{R}^{{n'}}}$, ${{\mathbf{U}}_s} \in {\mathbb{R}^{{n'} \times n}}$, ${{\mathbf{U}}_a} \in {\mathbb{R}^{{n'} \times C}}$ and ${{\mathbf{U}}_{f}} \in {\mathbb{R}^{{n'} \times q}}$. The high-resolution context vector ${\mathbf{cB}_t}$ is computed based on another coverage based attention model $f_{\text{catt}}$ with different initialized parameters except the ${{\mathbf{U}}_s}$ transition matrix.

\section{Training and Testing Details}
\label{sec:Training and Testing Details}
We validated the proposed model on CROHME 2014~\cite{mouchere2014icfhr} test set and CROHME 2016~\cite{mouchere2016icfhr2016} test set. The CROHME competition dataset is currently the most widely used public dataset for handwritten mathematical expression recognition. The training set has 8,836 expressions including 101 math symbol classes. The CROHME 2014 test set has 986 expressions while the CROHME 2016 test set has 1,147 expressions.

\subsection{Training}
\label{sec:Training}
The training objective of the proposed model is to maximize the predicted symbol probability as shown in Eq. \eqref{eq:computePy} and we use cross-entropy (CE) as the objective function:
\begin{equation}\label{eq:objective}
  O = - \sum\nolimits_{t=1}^T \log p({w_t}|{\mathbf{y}_{t-1},\mathbf{x}})
\end{equation}
where $w_t$ represents the ground truth word at time step $t$.

The implementation details of Dense encoder has been introduced in Section~\ref{sec:Multi-Scale Dense Encoder}. The decoder is a single layer with 256 forward GRU units. The embedding dimension $m$ and decoder state dimension $n$ are set to 256. The multi-scale attention dimension $n'$ is set to 512. The convolution kernel size for computing low-resolution coverage vector is set to $11 \times 11$ but $7 \times 7$ for high-resolution coverage vector, while their number of convolution filters are both set to 256. We utilized the adadelta algorithm \cite{zeiler2012adadelta} with gradient clipping for optimization. The best model is determined in terms of word error rate (WER) of validation set. We used a weight decay of ${10^{ - 4}}$ and we added a dropout layer~\cite{srivastava2014dropout} after each convolutional layer and set the drop rate to 0.2.

\subsection{Decoding}
\label{sec:Decoding}
In the decoding stage, we aim to generate a most likely LaTeX string given the input image. However, different from the training procedure, we do not have the ground truth of previous predicted symbol. Consequently, a simple left-to-right beam search algorithm~\cite{cho2015natural} is employed to implement the decoding procedure. Here, we maintained a set of 10 partial hypotheses at each time step, ending with the end-of-sentence token $<eos>$. We also adopted the ensemble method \cite{dietterich2000ensemble} for improving the performance. We first trained 5 models on the same training set but with different initialized parameters. Then we averaged their prediction probabilities during the beam search process.

\section{Experiments}
\label{sec:Experiments}
We designed a set of experiments to validate the effectiveness of the proposed method for handwritten mathematical expression recognition by answering the following questions:
\begin{description}
    \item[Q1] Is the dense encoder effective?
    \item[Q2] Is the multi-scale attention model effective?
    \item[Q3] Does the proposed approach outperform state-of-the-arts?
\end{description}

\subsection{Metric}
\label{sec:Metric}
The participating systems in all of the CROHME competitions were ranked by expression recognition rates (ExpRate), i.e., the percentage of predicted mathematical expressions matching the ground truth, which is simple to understand and provides a useful global performance metric. The CROHME competition compared the competing systems not only by ExpRate but also those with at most one to three symbol-level errors. In our experiments, we first transferred the generated LaTeX strings into MathML representation and then computed these metrics by using the official tool provided by the organizer of CROHME. However, it seems inappropriate to evaluate an expression recognition system only at expression level. So we also evaluated our system at symbol-level by using WER metric.

\subsection{Evaluation of dense encoder (Q1)}
\label{sec:Evaluation of dense encoder (Q1)}
We start the proposed multi-scale attention model with dense encoder from WAP~\cite{zhang2017watch}. As shown in Table~\ref{tab:1}, WAP achieves an ExpRate of 44.4\% on CROHME 2014 test set and an ExpRate of 42.0\% on CROHME 2016 test set. The WAP employs an encoder based on VGG architecture and its decoder is a unidirectional GRU equipped with coverage based attention model. Here, we only replace the VGG encoder by dense encoder with the other settings keeping unchanged and the new system is named as ``Dense'' in Table~\ref{tab:1}. The implementation details of the dense encoder is illustrated by the main branch in Fig.~\ref{fig:multi-scale-dense}. We can observe that the ExpRate increases about 5.7\% on CROHME 2014 and 5.5\% on CROHME 2016 by employing dense encoder.

\begin{table}[h]
\caption{\label{tab:1}{Comparison of recognition performance (in \%) on CROHME 2014 and CROHME 2016 when employing dense encoder and multi-scale attention model.}}
\centering
\begin{tabular}{c c c c c}
\toprule
\multirow{2}{*}{System} & \multicolumn{2}{c}{CROHME 2014} & \multicolumn{2}{c}{CROHME 2016} \\
\cmidrule(lr){2-3}
\cmidrule(lr){4-5}
 & WER & ExpRate & WER & ExpRate\\
\midrule
\textbf{WAP} & 19.4 & 44.4 & 19.7 & 42.0 \\
\textbf{Dense} & 13.9 & 50.1 & 15.4 & 47.5 \\
\textbf{Dense+MSA} & 12.9 & 52.8 & 13.7 & 50.1 \\
\bottomrule
\end{tabular}
\end{table}

\subsection{Evaluation of multi-scale attention model (Q2)}
\label{sec:Evaluation of multi-scale attention model (Q2)}
In Table~\ref{tab:1}, the system ``Dense+MSA'' is the proposed multi-scale attention model with dense encoder. ``+MSA'' means that we only replace the single-scale coverage based attention model in system ``Dense'' by multi-scale coverage based attention model. The performance of multi-scale attention model is clear to be observed by the comparison between system ``Dense'' and system ``Dense+MSA''. The ExpRate increases from 50.1\% to 52.8\% on CROHME 2014 and from 47.5\% to 50.1\% on CROHME 2016 after the implementation of multi-scale attention model.

More specifically, in the system ``Dense+MSA'', the multi-scale branch of dense encoder contains a dense block with depth $D=16$. We choose $D=16$ as we investigate the depth of block in multi-scale branch ($D=0, 8, 16, 24$) by experiments. In Table~\ref{tab:2}, $D=0$ means that we simply choose the output of the last transition convolutional layer in the main branch of dense encoder as the high-resolution annotations. The performance is only slightly improved compared with system ``Dense'' in Table~\ref{tab:1} which implies that more convolution operations are necessary to obtain more semantic high-resolution annotations. We can observe that $D=16$ is the best setting for both test sets of CROHME 2014 and 2016. The unpleasant results of $D=24$ indicate that too many convolution operations in the multi-scale branch can also lead to performance degradation.

\begin{table}[h]
\caption{\label{tab:2}{Comparison of recognition performance (in \%) on CROHME 2014 and CROHME 2016 when increasing the depth of dense block in multi-scale branch.}}
\centering
\begin{tabular}{c c c c c}
\toprule
\multirow{2}{*}{Depth} & \multicolumn{2}{c}{CROHME 2014} & \multicolumn{2}{c}{CROHME 2016} \\
\cmidrule(lr){2-3}
\cmidrule(lr){4-5}
 & WER & ExpRate & WER & ExpRate\\
\midrule
\textbf{0} & 13.5 & 50.8 & 14.3 & 48.5 \\
\textbf{8} & 13.3 & 51.3 & 13.9 & 49.3 \\
\textbf{16} & 12.9 & 52.8 & 13.7 & 50.1 \\
\textbf{24} & 13.1 & 51.4 & 14.1 & 48.9 \\
\bottomrule
\end{tabular}
\end{table}

\begin{figure}
\centering
\includegraphics[width=2.5in]{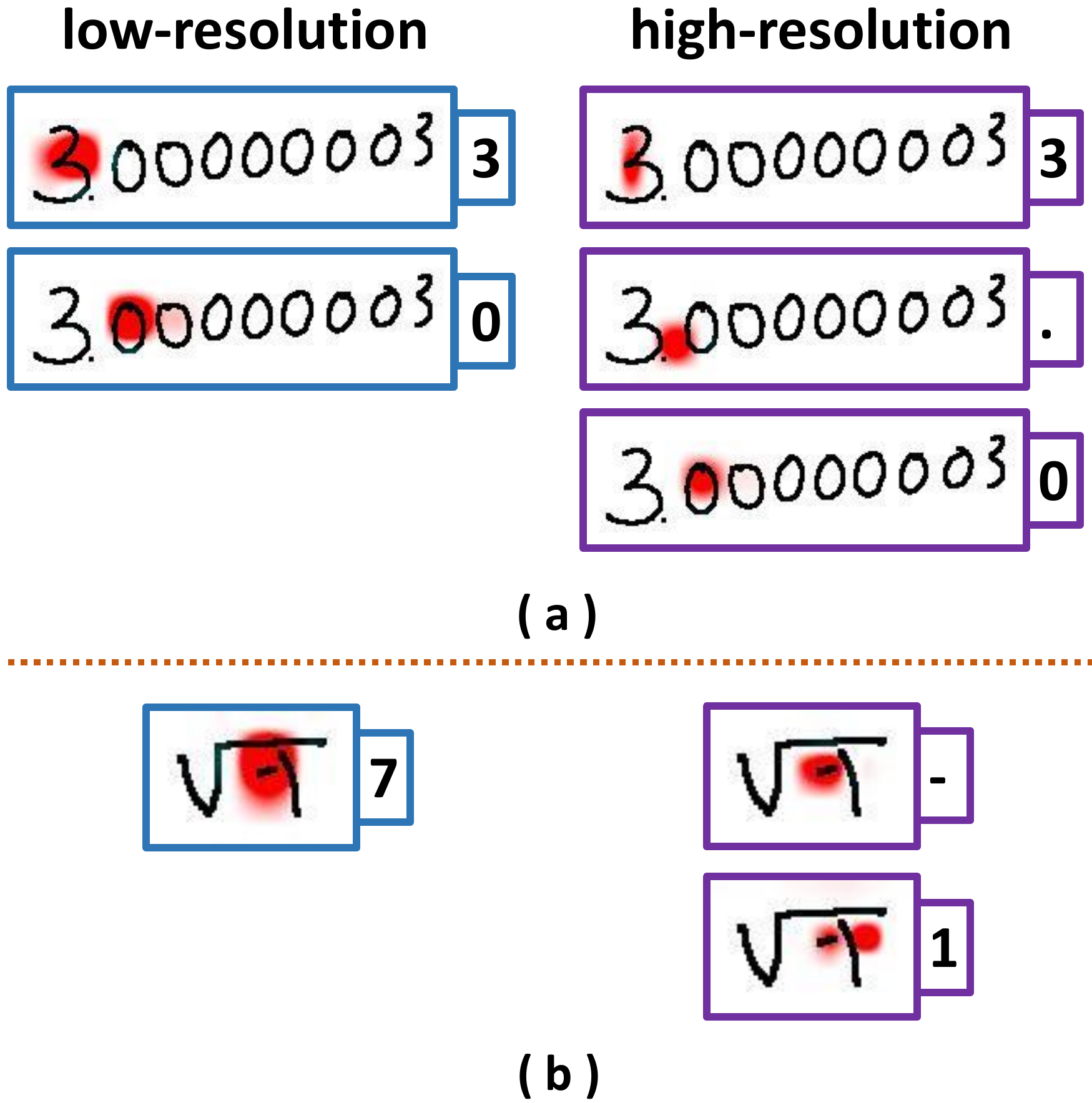}
\caption{Two examples of attention visualization on low-resolution annotations and on high-resolution annotations. The attention probabilities are shown through red color and the predicted symbols are shown on the right of images.}
\label{fig:attention_visualization}
\end{figure}

We also illustrate the performance of multi-scale attention model in Fig.~\ref{fig:attention_visualization}. The left part of Fig.~\ref{fig:attention_visualization} denotes the visualization of single-scale attention on low-resolution annotations and the right part denotes the visualization of multi-scale attention only on high-resolution annotations. Fig.~\ref{fig:attention_visualization} (a) is an example that the decimal point ``.'' is under-parsed by only relying on low-resolution attention model. However, the high-resolution attention in the multi-scale attention model successfully detects the decimal point. Fig.~\ref{fig:attention_visualization} (b) is an example that the math symbols ``- 1'' are mis-parsed as ``7'' due to the low-resolution attention model while the high-resolution attention model can correctly recognize them.

\subsection{Comparison with state-of-the-arts (Q3)}
\label{sec:Comparison with state-of-the-arts (Q3)}

\begin{table}[h]
\caption{\label{tab:3}{Comparison of ExpRate (in \%) on CROHME 2014, we erase system \uppercase\expandafter{\romannumeral3} namely MyScript because it used extra training data.}}
\centering
\begin{tabular}{c c c c c}
\toprule
\textbf{System} & \textbf{Correct(\%)} & \textbf{$\leq$ 1(\%)} & \textbf{$\leq$ 2(\%)} & \textbf{$\leq$ 3(\%)}\\
\midrule
\uppercase\expandafter{\romannumeral1} & 37.2 & 44.2 & 47.3 & 50.2 \\
\uppercase\expandafter{\romannumeral2} & 15.0 & 22.3 & 26.6 & 27.7 \\
\uppercase\expandafter{\romannumeral4} & 19.0 & 28.2 & 32.4 & 33.4 \\
\uppercase\expandafter{\romannumeral5} & 19.0 & 26.4 & 30.8 & 33.0 \\
\uppercase\expandafter{\romannumeral6} & 25.7 & 33.2 & 35.9 & 37.3 \\
\uppercase\expandafter{\romannumeral7} & 26.1 & 33.9 & 38.5 & 40.0 \\
\midrule
\textbf{WAP} & \textbf{44.4} & \textbf{58.4} & \textbf{62.2} & \textbf{63.1} \\
\textbf{CRNN} & \textbf{35.2} & \textbf{-} & \textbf{-} & \textbf{-} \\
\textbf{Ours} & \textbf{52.8} & \textbf{68.1} & \textbf{72.0} & \textbf{72.7} \\
\bottomrule
\end{tabular}
\end{table}

The comparison among the proposed approach and others on CROHME 2014 test set is listed in Table~\ref{tab:3}. Systems \uppercase\expandafter{\romannumeral1} to \uppercase\expandafter{\romannumeral7} were submitted systems to CROHME 2014 competition and they were mostly based on the traditional two-dimensional context free grammar method. Details of these systems can refer to \cite{mouchere2014icfhr}. To make a fair comparison we erase the system \uppercase\expandafter{\romannumeral3} namely ``MyScript'' which achieved a high ExpRate of 62.7\% but used a large private dataset and the technical details were unrevealed. System ``WAP'', ``CRNN'' and our proposed system are all based on encoder-decoder model with attention that takes handwritten mathematical expressions input as images. As for the system ``CRNN'', it is declared in \cite{anh2017training} that the encoder employs a CNN+RNN architecture and the decoder is a unidirectional RNN with classic attention model. Meanwhile a novel data augmentation method for handwritten mathematical expression recognition was proposed in~\cite{anh2017training}. We can see that our proposed system achieves the best result with ExpRate of 52.8\% on CROHME 2014. Additionally, a gap existed between the correct and error percentages ($\leq$ 1\%), showing that the corresponding systems have a large room for further improvements. In contrast, the small differences between error ($\leq$ 2\%) and error ($\leq$ 3\%) illustrate that it is difficult to improve the ExpRate by incorporating a single correction.

\begin{table}[h]
\caption{\label{tab:CROHME2016}{Comparison of ExpRate (in \%) on CROHME 2016, we erase team MyScript because it used extra training data.}}
\centering
\begin{tabular}{c c c c c}
\toprule
\textbf{} & \textbf{Correct(\%)} & \textbf{$\leq$ 1 (\%)} & \textbf{$\leq$ 2 (\%)} & \textbf{$\leq$ 3 (\%)}\\
\midrule
Wiris & 49.6 & 60.4 & 64.7 & -- \\
Tokyo & 43.9 & 50.9 & 53.7 & -- \\
S$\widetilde{\textbf{a}}$o Paolo & 33.4 & 43.5 & 49.2 & -- \\
Nantes & 13.3 & 21.0 & 28.3 & -- \\
\midrule
\textbf{WAP} & \textbf{42.0} & \textbf{55.1} & \textbf{59.3} & \textbf{60.2} \\
\textbf{Ours} & \textbf{50.1} & \textbf{63.8} & \textbf{67.4} & \textbf{68.5} \\
\bottomrule
\end{tabular}
\end{table}
To complement a more recent algorithm comparison and test the generalization capability of our proposed approach, we also validate our best system on CROHME 2016 test set as shown in Table \ref{tab:CROHME2016}, with an ExpRate of 50.1\% which is quite a promising result compared with other participating systems. The system ``Wiris'' was awarded as the first place on CROHME 2016 competition using only the CROHME training data with an ExpRate of 49.6\%, and it used a Wikipedia formula corpus, consisting of more than 592,000 LaTeX notations of mathematical expressions, to train a strong language model. The details of other systems can be found in \cite{mouchere2016icfhr2016}.

\section{Conclusion}
\label{sec:Conclusion}
In this study we improve the performance of attention based encoder-decoder for handwritten mathematical expression by introducing the dense encoder and multi-scale attention model. It is the first work that employs densely connected convolutional networks for handwritten mathematical expression recognition and we propose the novel multi-scale attention model to alleviate the problem causing by pooling operation. We demonstrate through attention visualization and experiment results that the novel multi-scale attention model with dense encoder performs better than the state-of-the-art methods.


\bibliographystyle{IEEEtran}
\bibliography{refs}

\end{document}